\documentclass[runningheads]{llncs}
\usepackage{graphicx}
\usepackage{algorithm}
\usepackage{algpseudocode}% http://ctan.org/pkg/algorithmicx
\usepackage{amsmath}
\usepackage{relsize}
\usepackage{verbatim}
\usepackage{verbatim}
\usepackage{booktabs}
\usepackage{siunitx}
\usepackage{multirow}
\usepackage[utf8]{inputenc}
\usepackage[english]{babel}
\usepackage{mathtools}
\usepackage{hyperref}
\hypersetup{
	colorlinks=true,
	citecolor=blue,
	linkcolor=blue,
	filecolor=blue,      
	urlcolor=blue,
}

\usepackage{amsmath}
\usepackage{graphicx}
\usepackage{tikz}
\usepackage{tikz}
\usetikzlibrary{automata,positioning}
\usepackage{amsmath}
\usepackage{chngcntr}
\usepackage{wrapfig}
\usepackage{color,soul}
\usepackage{lipsum}
\usepackage{color, colortbl}
\definecolor{LRed}{rgb}{1,.8,.8}
\definecolor{MRed}{rgb}{1,.6,.6}
\definecolor{HRed}{rgb}{1,.2,.2}
\usepackage{float}
\begin{document}
\title{Inferring  linear and nonlinear  Interaction networks using neighborhood support vector machines\thanks{Supported by  "HERMES" project,  Erasmus Mundus European programme, action 2,  Marseilles, France.}}
\author{Kamel Jebreen\inst{1, 3}\and
Badih Ghattas\inst{2}}
\authorrunning{J. Kamel et al.}
\institute{Institut de Mathématiques de Marseille, CNRS, UMR 7373,  Aix Marseille UniversityMarseille, France.\\
\email{}\and
 Institut de Mathématiques de Marseille, CNRS, UMR 7373, Aix Marseille UniversityMarseille, France.
 \\ \email{}\and Department of Mathematics, An-Najah National University, Nablus, Palestine}
\maketitle              % typeset the header of the contribution
\begin{abstract}
In this paper, we consider modelling interaction between a set of variables in the context of time series and high dimension. We suggest two approaches. The first is similar to the neighborhood lasso when the lasso model is replaced by a support vector machine (SVMs). The second is a restricted Bayesian network adapted for time series. We show the efficiency of our approaches by simulations using linear, nonlinear data set and a mixture of both.

\end{abstract}

\section{Introduction}
Modelling interactions between variables is a common task in statistics. This is often done using graphical models (\cite{lauritzen_graphical_1996})  where vertices correspond to variables and edges to the interaction between the corresponding variables. Such models may be inferred from data using different approaches. 
Among these approaches covariance graphs are the simplest as they infer the network applying a threshold to the estimated correlation matrix (\cite{cox_multivariate_1996},\cite{butte_discovering_2000}) .  Graphical Gaussian models (GGMs) (\cite{whittaker_graphical_1990,dempster_covariance_1972}), (\cite{lauritzen_graphical_1996})  consider rather partial correlations obtained from the inverse of the covariance matrix (\cite{schafer_empirical_2005}, \cite{schafer_learning_2005}). 

Bayesian networks (BN) (\cite{friedman_learning_1998}) infer interactions by estimating the conditional independence between the variables based on a specific factorization of the joint probabilities of the variables.  
Recently, the neighborhood lasso (\cite{meinshausen_high_2006}) and graphical lasso (\cite{friedman_sparse_2008}) suggest fitting a regression model for each variable using the others. A variable is connected in the graph to the set of its explanatory variables whose coefficient in the regression model are not zero. 

These approaches have been extended to time series data. Dynamic Bayesian networks (DBN) (\cite{friedman_learning_1998}) are such direct extension of Bayesian networks. For the neighborhood lasso, a variable $X$ at time point $t+1$ is regressed on all the variables observed at the previous time point $t$.

In this work, we suggest first a similar approach to neighborhood lasso replacing the lasso model by SVM where the subset of variables used for each regression is obtained by a feature selection approach.  We experiment also a static restricted Bayesian network for time series data. 

This work is organized as follows.  Section 2 gives a brief introduction to graphical models for time series. Section 3 describes our approaches. Finally,  Section 4 is dedicated to simulations and results. 

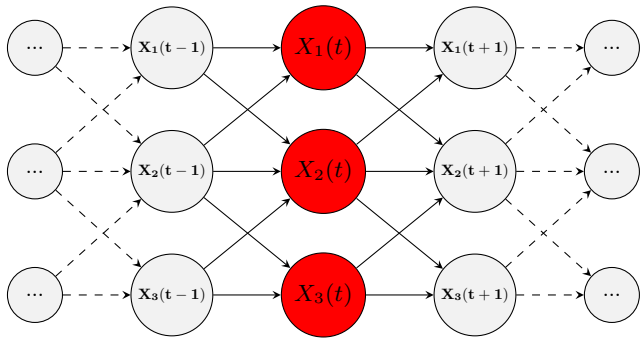
\begin{figure}
	\centering
	%\begin{tikzpicture}[scale=\tikzscale]
	\begin{tikzpicture}[>=stealth,state/.append style={fill=gray!10}]
	\begin{scope}[scale=0.9, transform shape]
	
	% vertices of autmaton 
	\node [state] 			(x11)								{$...$};
	%\node [draw=none] 		(x12)		[right=of x11]							{$...$};
	\node[state, scale=0.7] 			(x1t-1) 		[right=of x11]				{ $\mathbf{X_1(t-1)}$};
	\node[state,fill=red, scale=1.1] 			(x1t)		[right= of x1t-1 ]			{$X_1(t)$};
	\node[state, scale=0.7]			(x1t+1)		[right= of x1t]				{ $ \mathbf{X_1(t+1)}$};
	\node[state] (x1t+2) [right=of x1t+1] {$...$};
	\node[state] 			(x21)		[below= of x11]				{$...$};
	\node[state, scale=0.7]			(x2t-1)		[right = of x21]			{ $\mathbf{X_2(t-1)}$};
	\node[state,fill=red, scale=1.1] 			(x2t)		[right= of x2t-1]				{$X_2(t)$};
	\node[state, scale=0.7]			(x2t+1)		[right= of x2t]				{ $\mathbf{ X_2(t+1)}$};
	\node[state] (x2t+2) [below = of x1t+2] {$...$};
	
	\node[state]			(x31)		[below= of x21]				{$...$};
	\node[state, scale=0.7]			(x3t-1)		[right= of x31]				{$\mathbf{X_3(t-1)}$};
	\node[state,fill=red, scale=1.1]			(x3t)		[right= of x3t-1]			{$X_3(t)$};
	\node[state, scale=0.7]			(x3t+1)		[right= of x3t]				{ $\mathbf{ X_3(t+1)}$};
	\node [state] 	(x3t+2)		[below=of x2t+2] {$...$};
	
	% edges of
	%\draw[dashed,->] (x11) edge node {} (x1t-1);
	%\draw[dashed,->] (x1t-1) edge node {} (x1t);
	%\draw (x1t-1) edge[->] node {} (x1t);
	\path[dashed,->] (x11) edge node {} (x1t-1);
	\path[dashed,->] (x11) edge node {} (x2t-1);
	%\path[->,bend left=15] (x1t-1) edge node {} (x1t);
	\path[->] (x1t-1) edge node {} (x1t);
	\path[->] (x1t-1) edge node {} (x2t);
	\path[->] (x1t) edge node {} (x1t+1);
	\path[->] (x1t) edge node {} (x2t+1);
	\path[dashed,->] (x1t+1) edge node {} (x1t+2);
	\path[dashed,->] (x1t+1) edge node {} (x2t+2);
	
	\path[dashed,->] (x21) edge node {} (x1t-1);
	\path[dashed,->] (x21) edge node {} (x2t-1);
	\path[dashed,->] (x21) edge node {} (x3t-1);
	\path[->] (x2t-1) edge node {} (x1t);
	\path[->] (x2t-1) edge node {} (x2t);
	\path[->] (x2t-1) edge node {} (x3t);
	\path[->] (x2t) edge node {} (x2t+1);
	\path[->] (x2t) edge node {} (x1t+1);
	\path[->] (x2t) edge node {} (x3t+1);
	\path[dashed,->] (x2t+1) edge node {} (x1t+2);
	\path[dashed,->] (x2t+1) edge node {} (x2t+2);
	\path[dashed,->] (x2t+1) edge node {} (x3t+2);
	
	\path[dashed,->] (x31) edge node {} (x2t-1);
	\path[dashed,->] (x31) edge node {} (x3t-1);
	\path[->] (x3t-1) edge node {} (x2t);
	\path[->] (x3t-1) edge node {} (x3t);
	\path[->] (x3t) edge node {} (x3t+1);
	\path[->] (x3t) edge node {} (x2t+1);
	\path[dashed,->] (x3t+1) edge node {} (x2t+2);				
	\path[dashed,->] (x3t+1) edge node {} (x3t+2);
	
	%\path[->,bend left=15,dark] (x1t) edge node  {} (x1t+1);
	%\path[->,bend left=15] (x1t+1) edge node  {} ();
	\end{scope}
	\end{tikzpicture}
	\caption{Graphical representation of a time varying dynamic Bayesian network. } 
	\label{fig:dbnfig}
\end{figure}

\section{Graphical models for time series}\label{sec:Graphical_models}
In this section, we give a brief summary of recent works on graphical models for time series. Let  $\mathbf X(t)=(X_1(t),...,X_p(t))$ be a vectorial real p-dimensional Gaussian process observed at time $t=1,..., n$.  $\mathbf X(t)$ is assumed to follow a normal distribution $\mathcal{N}(\mu, \Sigma),$ where $\mu$ is the mean vector and $\Sigma$ is the covariance matrix. All the approaches described in this section make the following assumption:
Firstly, we assume that  $\mathbf X$ is first order Markovian, that is 
%$$X(t)\perp X(t'<t-1)|X(t-1), \;\; t\geq 3$$
\begin{equation}\label{eq:ass1}
P( X(t+1)| X(1),..., X(t))=P( X(t+1)| X(t))
\end{equation}
this means that the variables $\mathbf X(t+1)$  depend only  on the past variables $\mathbf X(t)$.
Secondly,  we assume  that the process is stationary, that is 
\begin{align}\label{eq:ass2}
P( X(t+1)| X(t))\;\; \text{is independent of $t$}.
\end{align}
Thirdly, the variables observed at a same time  are conditionally  independent given the others in the past time, that is
\begin{equation}\label{eq:ass3}
X_i(t) \perp X_j(t)|\mathbf X(t'<t) \;\;\textrm{ where} \;i\neq j, t, t' \geq 1.
\end{equation}

These assumptions ensure the existence of a directed acyclic graph (DAG) $G=(\mathbf X(t),E(G))$, where $\mathbf X(t)$ is the set of variables or nodes and $E(G)\subseteq (\mathbf X(t) \times \mathbf X(t))$ is the set of edges. Then, a  Bayesian network corresponds to the following  representation of the joint distribution of $\mathbf X$
\begin{equation}\label{eq:bndesnity}
f(\mathbf X(t))=\prod\limits_{j=1}^p\prod\limits_{t=1}^nf(X_j(t)|Pa(X_j(t),G)) 
\end{equation}
where $Pa(X_j(t),G)$ is the set of parents of $X_j(t)$ in the graph $G$.
\subsection{Dynamic Bayesian networks (DBN)}
Friedman \textit{et.al.} (\cite{friedman_learning_1998}) suggests two parts to model the process $\mathbf X(t)$ using DBN. The first is a  prior network $B_0$ that determines the distribution of the initial states $\mathbf X(1),$ and the second is a  transition network $B_{\rightarrow}$ which determines the transition probability $P(X(t+1)|X(t))$ for all $t$. That is
\begin{equation}\label{eq:freidmanbic}
P_{B_{\rightarrow}}(\mathbf X(1),...,\mathbf X(n))=P_{B_0}(\mathbf X(1))\prod\limits_{t=1}^{n-1}P_{B_{\rightarrow}}(\mathbf X(t+1)|\mathbf X(t)). 
\end{equation}
The structure of a DBN is optimized using Bayesian Information Criterion  ($BIC$) score  defined as follows

\begin{equation}
BIC(\mathbf X(t), G)=BIC_0+BIC_{\rightarrow B},
\end{equation}
such that,
\begin{align}\label{eq:BIC}
BIC(\mathbf X(t), G)=\sum\limits_{ j =1}^p \log \hat{f}(X_j(t+1)|X_j(t))-\frac{L}{2}\log n,
\end{align}
where $BIC_0$ is the  $BIC$ score of the prior network $B_0,$ $BIC_{\rightarrow B}$ is the $BIC$ score for the transition network $B_{\rightarrow},$ $L$ is the number of parameters in $G$  and  $\hat{f}(X_j(t+1)|X_j(t))$ is the local conditional  distribution for each variable.

The next approaches are based on the covariance matrix estimation.
%\subsection{Approaches based on the covariance matrix estimation}
\subsection{Least Angle Absolute Shrinkage and Selection Operator (Lasso)}
% ,   where $\mathbf{X_{-j}}$ are all variables except $\mathbf X_j$, $j=1,...,p$ 
Meinshausen and Bühlmann (\cite{meinshausen_high_2006}) used the Lasso  approach (\cite{tibshirani_regression_1994}) for inferring the concentration matrix which is the inverse covariance matrix. They apply the lasso regression for each variable as a response variable given the others. That is, fit the variable $X_j(t+1)$ at time point $t+1$ over all variables $\mathbf X(t)$ at the previous time point $t$ for all $j=1,...,p$, and the coefficients of the regression are given by 
\begin{align}
\hat{\beta}^{j,\lambda}=\underset{\beta:\beta_j=0}{argmin}\left[ \frac{\parallel X_j(t+1)-\beta \mathbf X(t)\parallel_2^2}{n}+\lambda \parallel\beta\parallel_1\right].
\end{align}

The models regressing $X_i$ over $X_j$ and $ X_j$ over $X_i$ may have zero or nonzero coefficients giving opposite information about the correlation of these variables. This is the symmetrization problem solved by applying "And" or "or" operations over such connections.

Friedman \textit{et.al.} (\cite{friedman_sparse_2008}) suggest the graphical lasso model regression, \textit{Glasso,}  as improvement of the neighborhood lasso reducing the cost of the computation mainly for high dimensional problems.

\subsection{Shrinkage approach (Genenet)}
Making  the assumption that $\mathbf X(t)$ follows a VAR process, 
Schäfer and Strimmer (\cite{schafer_shrinkage_2005}) suggest to estimate the partial correlation matrix using a  James Stein type shrinkage  estimator (\cite{efron_steins_1973}). This estimator is much more efficient when compared to least square or maximum likelihood estimates mainly when the sample size  is lower than the dimension; $n<p.$

\subsection{First order conditional dependence graph (G1DBN)}
Lèbre (\cite{lebre_inferring_2009}) proposed an approach which proceeds using   two steps. The first aims to estimate the adjacency matrix with a reduced number of edges. Conditional Partial correlation of variables $X_i$ and $X_j$ over $X_k$, $i,j, k=1,...,p$ where  $k\neq j$,  are computed together with their \textit{p-values} $p_{ij|k}$; the maximal over $k$ is kept giving a score matrix $S_{ij}=\underset{k}{max}\; p_{ij|k}.$ A first graph is obtained applying a threshold $\alpha_1$ to these scores. 
The second step changes the score matrix using \textit{p-values} of significance testing of linear regression coefficients over a restricted subset of variables (those whose initial score are nonzero).
\subsection{ Statistical Inference for Modular Networks, SIMoNe}
Ambroise \textit{et.al.} ()\cite{ambroise_inferring_2009}) suggest an algorithm called \textit{SIMoNe} (Statistical Inference for Modular Networks) to estimate the nonzero entries of the concentration matrix which is equivalent to reconstructing the Gaussian graphical model.  They assume a latent structure on the concentration matrix which is equivalent to a hidden structure over the network (whose edges weighs correspond to the entries of the concentration matrix). Finally, they use an EM algorithm together with a $\ell_1$ norm to get the concentration matrix estimate.

Other approaches based in mutual information criterion  were also proposed, for details see (\cite{altay_inferring_2010,basso_reverse_2005,faith_large_scale_2007,meyer_information_theoretic_2007,peng_feature_2005,reverter_combining_2008}).

\section{Our approaches}\label{sec:our_approches}
We propose two new approaches for inferring dynamic graphical networks: neighborhood SVM (nSVM) and restricted Bayesian networks (RBN). 
Support vector machines (\cite{vapnik_nature_1995}) may be used for regression and share many features with the classification version.

Suppose we have training data set $D=\lbrace ( x_1,y_1),...,( x_n,y_n) \rbrace \subset \mathcal R^p \times \mathcal R,\; p\geq 1$. In epsilon SVM regression ($\epsilon-SVM$)  we aim to estimate a function $f(x)$ that has at most $\epsilon$ deviation from the actual  targets $y_i$ for all the training data. Let 
\begin{align}\label{eq:linearSVM}
f(x)=<w,x>+b, \;\;\;\;\; w\in \mathcal R^p, b\in \mathcal R,
\end{align}
%from  eq (\ref{linear}) we are searching for a small value of $w$, i.e., 
$\epsilon-$SVM solve the following optimization problem:

%\begin{align}
%\text{subject to} 
\begin{equation}\label{eq:optimization}
\begin{aligned}
\text{minimize}&\;\;\;\;\frac{1}{2}\parallel w \parallel^2+C\sum\limits_{i=1}^n(\zeta_i+\zeta_i^*),\\
\text{subject to}&
\begin{cases}
y_i-<w, x_i>-b \leq \epsilon+\zeta_i,\\
<w, x_i>+b-y_i \leq \epsilon+\zeta_i^*,\\
\zeta_i, \zeta_i^* \geq 0,
\end{cases},
\end{aligned}
\end{equation}
%is trade off  constant between $f$ and the amount larger than $\epsilon$
where $C >0$ is a constant that controls the penalty imposed on observations which lie outside the $\epsilon$ margin and prevent overfitting and $\zeta_i, \zeta_i^*$ are slack variables controlling the relaxation of the constraints. 
The linear $\epsilon$-insensitive loss function ignores errors that are within $\epsilon$ distance of the observed value by treating them as equal to zero. The loss function is a measure based on the distance between observed value $y$ and the $\epsilon$ boundary:
\begin{align}
L_{\epsilon}=\begin{cases}
0, & |y-f(x)|\leq \epsilon\\
|y-f(x)|-\epsilon, & otherwise
\end{cases}.
\end{align}

Solving the optimization problem (\ref{eq:optimization}) is done by solving its Lagrange dual formulation (\cite{fletcher_practical_1987,mangasarian_nonlinear_1969,mccormick_nonlinear_1983})  and it  gives: 
\begin{align} \label{eq:weight}
w&=\sum\limits_{i=1}^n(\alpha_i-\alpha_i^*)x_i,\\ 
f(x)&=\sum\limits_{i=1}^n(\alpha_i-\alpha_i^*)<x_i,x>+b,
\end{align}
where $\alpha_i$ and $\alpha_i^*$ are the Lagrange multipliers. The constant $b$ can be computed by the so called Karush Kuhn Tucker (KKT) conditions (\cite{kuhn_nonlinear_1951,karush_minima_1939}).

So, $w$ can be completely described as a linear combination of the training patterns $x_i$ and the complexity of a function's representation by support vectors.  Also, it depends on the number of support vectors but not on the dimension $p$. 

In case of nonlinear SVMs the Lagrange dual formulation is  extended to a nonlinear function. Nonlinear SVM regression model can be obtained by replacing the dot product term $<x_j, x>=x_j^Tx$ with a nonlinear kernel function $K(x_1, x_2)=<\phi(x_1),\phi(x_2)>$, where $\phi(x) $ is a transformation that maps $x$ to a high dimensional space. The common kernels are 

\begin{align} \label{eq:kernal_svm}
K(x_i, x_j)=
\begin{cases}
\text{Polynomial kerel}=(k^* <x_i, x_j>+const)^d,\\
\text{Gaussian radial basis function}=e^{-k^*\parallel x_i-x_j \parallel^2}, \\
\text{Sigmod kernel}= tanh(k^* <x_i, x_j>+const),\\
\end{cases}
\end{align}
where $k^*$ is the kernel parameter, $d$ is the degree of the polynomial kernel and $const$ is a random constant.
\begin{figure}[H]
		\centering
		\includegraphics[scale=0.8]{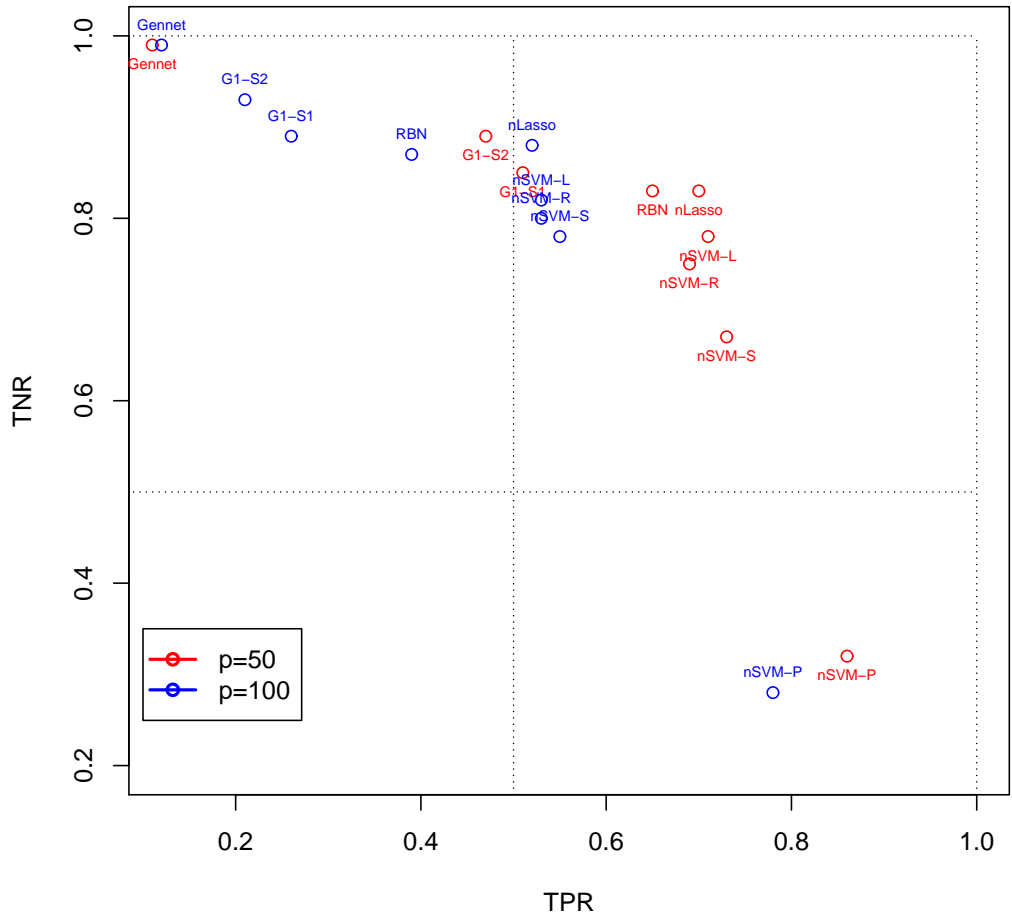}
		\caption{ Linear simulation:  position of approaches  with respect to TPR and TNR for $p=50,100$ and  $n=20$.} 
		\label{fig:VAR}
\end{figure}

\begin{figure}[H]
	\centering
	\includegraphics[scale=0.8]{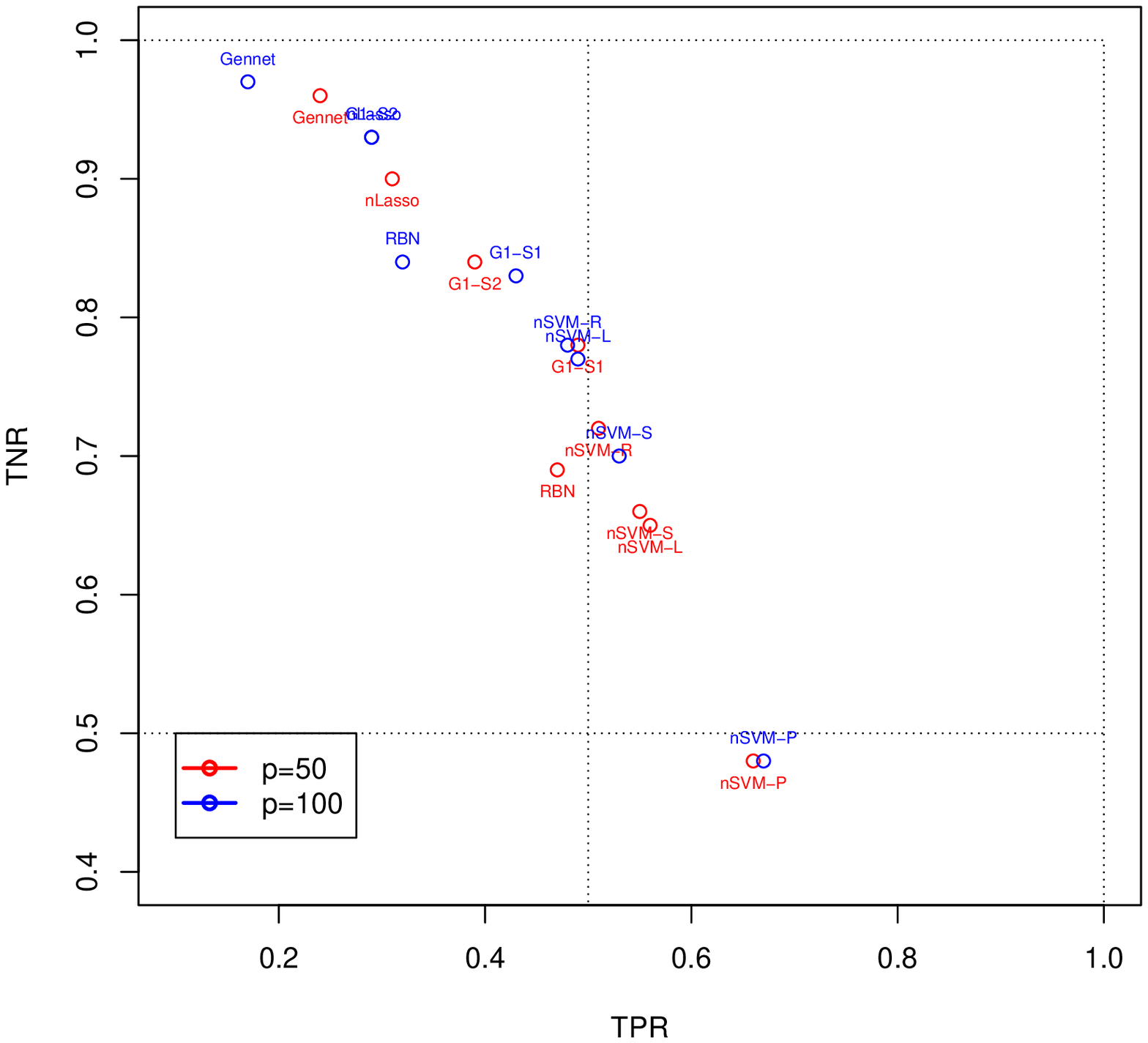}
	\caption{  Nonlinear simulation:  position of approaches  with respect to TPR and TNR for $p=50,100$ and  $n=20$.} 
	\label{fig:nonVAR}
\end{figure}

\begin{figure}[H]
	\includegraphics[scale=0.8]{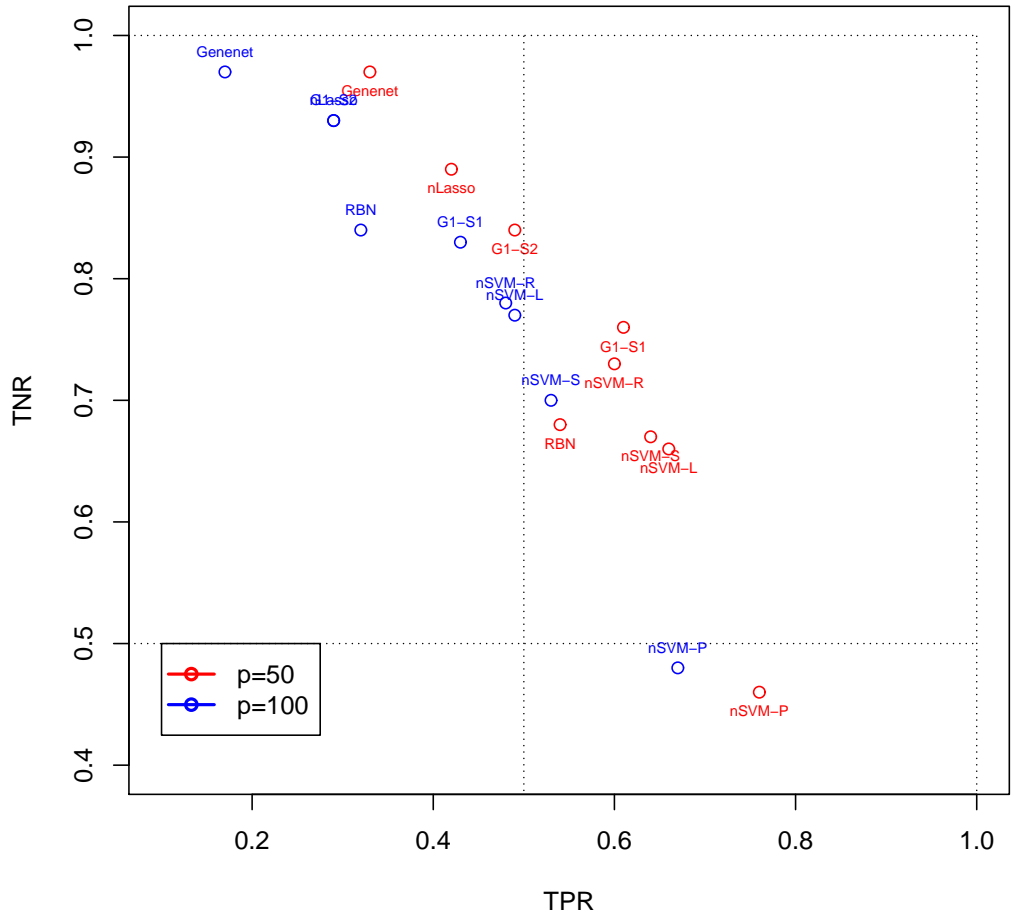}
	\caption{  Mixed  simulation:  position of approaches with respect to TPR and TNR for $p=50,100$ and  $n=20$.} 
	\label{fig:VARnonVAR}
\end{figure}

%\begin{figure*}[!htp]
%	\begin{subfigure}[b]{0.5\textwidth}
%		\centering
%		\includegraphics[scale=0.5]{VAR.eps}
%		\caption{\centering Linear simulation:  position of approaches  with respect to TPR and TNR for $p=50,100$ and  $n=20$.} 
%		\label{fig:VAR}
%	\end{subfigure}
%	\hfill
%	\begin{subfigure}[b]{0.5\textwidth}
%		\centering
%		\includegraphics[scale=0.5]{nonVAR.eps}
%		\caption{\centering  Nonlinear simulation:  position of approaches  with respect to TPR and TNR for $p=50,100$ and  $n=20$.} 
%		\label{fig:nonVAR}
%	\end{subfigure}
%	\hfill
%	\begin{subfigure}[b]{0.5\textwidth}
%		\includegraphics[scale=0.5]{VARnonVAR.eps}
%		\caption{\centering  Mixed  simulation:  position of approaches with respect to TPR and TNR for $p=50,100$ and  $n=20$.} 
%		\label{fig:VARnonVAR}
%	\end{subfigure}
%	\caption{Approaches positions in linear, nonlinear simulation, and mixture of both for   $p=50, 100$ and  $n=20$. }
%	\label{fig:approach position}
%\end{figure*}
%% \blindtext

\subsection{Neighborhood Support vector machine, nSVM}
Following the idea of neighborhood lasso we suggest here a procedure based on $p$ SVM regression models where each variable observed at time $t+1$ plays the role of the output and the other variables observed at time $t$ are used as input variables. The difference with neighborhood lasso is that feature selection step is done separately. For each regression model, the optimal subset of input variables to keep is selected by a stepwise kind procedure. First, input variables are ranked according to their decreasing order of importance. The importance of variable $X_j$ based on SVM is computed using  $||w||^{(-j)},$ (\cite{cristianini_introduction_2000,rakotomamonjy_variable_2003,yu_svm_2012,lin_feature})  which is the norm of the weight vector omitting its $jth$ coordinate.
Once the input variables are ordered, we construct a sequence of embedded models beginning with the most important variable, and adding the others one by one (\cite{ben_ishak_2007}). The mean square error (MSE) of each model is computed by leave one out cross validation (LOOCV). The model minimizing the MSE corresponds to the best subset selection of input variables, thus the optimal neighbor. The algorithm  is summarized in algorithm  \ref{algo:nSVM_algorthim}.

\begin{algorithm}[H]
	\caption{Neighborhood Support vector machine algorithm.}
	\begin{algorithmic}[1]	
		%\begin{description}
		\State Let $D$ be a data set; $p$ is the number of features, $error$ and $Error$ be  vectors of MSE with length $p$ ;
		\For{\texttt ( j=1:p)}
		%from $D$ using permutation importance of random forests.\\
		\State Build a SVM model $f$ for each response variable $X_j(t+1)$ and predictor variables $\mathbf{X}(t)$;
		\State compute the variable importance (VI) with respect to the SVM model;
		\State Sort the variables according to their descending order of importance: $X^{(1)}(t),...,X^{(p)}(t)$;
		\State Partition $D$ using LOOCV  and let $D_{-i}=D\backslash D_i$;
		\State Initialize $Error=0$
		
		\For { \texttt ( i=1:n) }
		
		\For {( \texttt k=1:p)}
		\State $M_i^k=f(response=X_j(t+1), predictors=X^{(1)}(t),...,X^{(k)}(t),D_{-i});$ 
		\State $error_i^k=Test(M_i^k, D_{i});$
		\EndFor
		\State  $Error=Error+error_i;$
		\EndFor
		\State $Error=\frac{1}{n}Error;$
		%\State  %$Error^i=\frac{1}{n}\sum\limits_{j=1}^{n}Error_j^k;$
		\State  $kopt=\underset{k}{argmin} \lbrace Error \rbrace$, where  $kopt$ is the optimal number of important variables to keep in the model.
		%\end{description}
		\EndFor
	\end{algorithmic}
	%\textbf{END}
	\label{algo:nSVM_algorthim}
\end{algorithm}

\subsection{Restricted Bayesian Networks (RBN)}
Our idea here is to use the classical static Bayesian network approach augmenting the data set by adding one time shift for each variable. Thus we built a Bayesian network over the set of $2p$  variables $(X_1(t+1), ..., X_p(t+1), X_1(t), ..., X_p(t))$  constraining the network to satisfy the assumptions given in  section \ref{sec:Graphical_models}. Figure  \ref{fig:dbnfig} illustrates these restrictions. There are no dependencies within each time and arcs between times are only in one direction $(t+1\rightarrow t).$  Besides, the graph is acyclic.

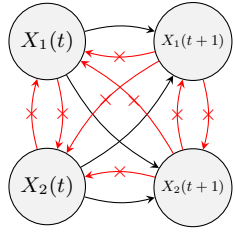
\begin{figure}[H]
	\centering
	%\begin{tikzpicture}[scale=\tikzscale]
	\begin{tikzpicture}[>=stealth,state/.append style={fill=gray!10}]
	\begin{scope}[scale=0.9, transform shape]
	\node [state] 		(x1t-1)								{  $X_1(t)$};
	\node [state, scale=0.7] 		(x1t)			[right= of x1t-1]	{ $X_1(t+1)$};
	\node [state] 		(x2t-1)			[below= of x1t-1]	{  $X_2(t)$};
	\node [state, scale=0.7] 		(x2t)			[below= of x1t]	{ $X_2(t+1)$};
	
	\path[->,bend left=15] (x1t-1) edge node {} (x1t);
	\path[->, bend right=15] (x1t-1) edge node {} (x2t);
	\path[->,  bend right=15] (x2t-1) edge node {} (x2t);
	\path[->, bend right=15] (x2t-1) edge node {} (x1t);
	
	\path[<-,bend right=15, red] (x1t-1) edge node {$\times$} (x1t);
	\path[<-,bend left=15, red] (x1t-1) edge node {$\times$} (x2t);
	\path[<-,  bend left=15,red] (x2t-1) edge node {$\times$} (x2t);
	\path[<-,bend left=15, red] (x2t-1) edge node {$\times$} (x1t);
	
	\path[->,bend left=15, red] (x1t) edge node {$\times$} (x2t);
	\path[<-,bend right=15, red] (x1t) edge node {$\times$} (x2t);
	
	\path[->,bend left=15, red] (x1t-1) edge node {$\times$} (x2t-1);
	\path[<-,bend right=15, red] (x1t-1) edge node {$\times$} (x2t-1);
	
	\end{scope}
	\end{tikzpicture}
	\caption{Restricted dynamic Bayesian network, RDBN.} 
	\label{fig:RDBN}
\end{figure}

\section{Experiments}
In this section we suggest three different simulation models for linear time series, nonlinear and a mixture of both. 
\subsection{Simulation models}
For the linear time series simulation we use a first order  vector autoregressive model (VAR(1))
\begin{align}
X_j(t+1)=A\mathbf{X(t)}+B+\epsilon_j, \; j=1, ... , p, 
\end{align}
where, $t=1, .., n$, $\mathbf{ X(t)} \in \mathcal{R}^p$ and $\epsilon_j \sim \mathcal{N}(0, \sigma^2)$. The matrix $A_{p\times p}$ represents the true network structure. Its elements are chosen uniformly fixing the true edges proportion $pi \in (0, 1)$ of non zeros entries.
The vector $B$ of intercepts is also chosen uniformly. Details are given in algorithm  \ref{algo:simuVAR},  (\cite{lebre_inferring_2009,opgen_rhein_learning_2007}).

For nonlinear time series, we follow the simulation scheme given  in (\cite{fujita_modeling_2008}) and use the following transformations
\begin{align} \label{eq:nonVarTrans}
%\textit{Nonlinear =}
\begin{cases}
f_1(X(t+1))=\sin( X(t))\\
f_2(X(t+1))=\cos( X(t))\\
f_3(X(t+1))=\sqrt[3]{ X^2(t)}-2^{\sin ( X(t))}
\end{cases}.
\end{align}
The initial values $ X(1)\in \mathcal{R}^p$ are drawn randomly using standard Gaussian distribution with zero mean and variance $\sigma^2.$ 

The $p$ nonlinear functions are drawn randomly from the above transformations (equation \ref{eq:nonVarTrans}) and applied to each dimension of $\mathbf{X}$ at time $t$. A matrix $A$ generated like in the linear case is applied after the nonlinear transformation. The process is described in algorithm \ref{algo:simuNonlineare}. 

To mix the linear and nonlinear simulation we use this set of  transformations
\begin{align} \label{eq:VARnonVarTrans}
%\textit{Linear /Nonlinear =}
\begin{cases}
f_4(X(t+1))=\sin( X(t))\\
f_5(X(t+1))=\frac{1}{2}  X(t)\\
f_6(X(t+1))=\sqrt[3]{ X^2(t)}-2^{\sin ( X(t))}\\
f_7(X(t+1))=-0.8 X(t)\\
\end{cases},
\end{align}
and we proceed exactly like for the nonlinear case.

To  test the  performance of our approaches  we compared the efficiency of neighborhood support vector machines approach (nSVM) and restricted Bayesian networks approach (RBN) with first order dependencies approach (G1DBN) (\cite{lebre_inferring_2009}), shrinkage to large scale covariance matrix estimation approach (\cite{schafer_shrinkage_2005,opgen_rhein_learning_2007}) (Genenet) and neighborhood lasso approach (nlasso) (\cite{meinshausen_high_2006}). Support vector machines depend on two  parameters  $k^*$ and $C=cost$ which are the kernel parameter and the constant of  regularization  in the Lagrange formulation respectively. These parameters are tuned and compared with their default values, $k^*=1/p$ and $C=1$ to choose the best performance (\cite{karatzoglou_support_2006}). The range of $k^*$  and $C$ are chosen respectively to be from $10^{-6}$ to $10^{-1}$  and from  $10^{1}$ from $10^{6}$.  Also, for polynomial kernel, the parameter $d$ is tuned to choose the best degree within $d=\lbrace 1,..,5\rbrace$. 

\begin{algorithm}[h]
	\caption{Simulation of linear networks}
	\begin{algorithmic}[1]	
		%\begin{description}
		\State Let ${X}={ Zero}_{p\times n}$ be the $p$-dimensional  time series   initialized to zero; $n$ the number of instances; $pi \in (0,1)$  the true edges proportion; $A={ Zero}_{p\times p}$  the adjacency matrix that represents the simulated graph  (initialized to zero); $B$  the intercept term; $\epsilon_i$ a white noise  with zero mean  and variance equal to $\sigma^2$; and  nEdges  the number of nonzero edges in the network;
		
		\State  nEdges=$ \lfloor p^2\times pi\rfloor$;
		\State Select  randomly the nonzero edges in $A$  from $p^2$ edges;
		\State Fill the nonzero edges in $A=<a_{ij}>$ uniformly;
		\State Define the true network  \begin{align}Tnet=
		\begin{cases}
		1, if \;\; a_{ij}\neq 0\\
		0, if \;\; a_{ij}= 0
		\end{cases};	\end{align}
		\State Draw the intercept term $B$ and the variance $\sigma^2$ uniformly;
		\State Draw  the initial value  $ X[,1]$ normally with zero mean  and variance equal to $\sigma^2$;
		\For {(\texttt i=2:n)}
		\Statex \hspace{0.2 in}$ X[,i]=A  X[,i-1]+B+\epsilon_i,;\;\;\;\;\;\; where\;\;\; \epsilon_i\sim \mathcal{N}(0, \sigma^2)$
		\EndFor
		\State $ X^T$ is the simulated times series.
	\end{algorithmic}
	%\textbf{END}
	\label{algo:simuVAR}
\end{algorithm}

\begin{algorithm}[h]
	\caption{Simulation of nonlinear and mixture networks}
	\begin{algorithmic}[1]	
		\State  Let ${X}={ Zero}_{p\times n}$ be the $p$-dimensional time series   initialized to zero ;  $n$  the number of instances; $pi \in (0,1)$  the true edges proportion; $A={ Zero}_{p\times p}$  the adjacency matrix that represents the simulated graph ( initialized to zero); $\epsilon_i$  a white noise with zero mean  and variance equal to $\sigma^2$; and  nEdges  the number of nonzero edges in the network;
		
		\State  nEdges=$\lfloor p^2\times pi \rfloor;$
		\State Select  randomly the nonzero edges in $A$  from $p^2$ edges;
		\State Fill the nonzero edges in $A=<a_{ij}>$ uniformly;
		\State Define the true network
		\begin{align}Tnet=
		\begin{cases}
		1, if \;\; a_{ij}\neq 0\\
		0, if \;\; a_{ij}= 0
		\end{cases};	\end{align}
		\State Choose  the transformation function $f_j$ randomly from equation \ref{eq:nonVarTrans} or equation \ref{eq:VARnonVarTrans} for each variable;
		\State Draw the initial value $ X[,1]$ normally with zero mean and random variance and set $ X[,1]=2\times \sin({X[,1]});$
		\For {( \texttt i=2:n)}
		\For {( \texttt j=1:p)}
		\Statex \hspace{0.2in} \;\;$X[j,i]=f_j(X[j,i-1])$
		\EndFor
		\Statex \hspace{0.2in}$\;\; X[,i]=A\times  X[,i]+\epsilon_i;\;\;\;\;\;\; where\;\;\; \epsilon_i\sim(0, \sigma^2)$
		\EndFor
		\State ${X}^T$ is the simulated times series.
	\end{algorithmic}
	%\textbf{END}
	\label{algo:simuNonlineare}
\end{algorithm}

\subsection{Results}
We compare  the different approaches described above; $G1 (S1$ and $S2)$ which correspond to the two steps of G1DBN approach, neighborhood lasso (nlasso) approach, the Genenet approach   with our approaches Restricted Bayesian network approach (RBN) and neighborhood SVM approach with different kernels; linear (L),  radial (R), sigmod (S),  and polynomial (P).

For these comparisons we compute the true positive rate (TPR), false positive rate(FPR), true negative rate  (TNR) and false negative rate (FNR) defined in equation \ref{eq:TFPNR} and average their values over  100 runs.

\begin{equation}\label{eq:TFPNR}
\begin{aligned}
TPR&=\frac{TP}{TP+FN},&& FPR=\frac{FP}{FP+TN}\\
TNR&=\frac{TN}{TN+FP},&& FNR=\frac{FN}{FN+TP}
\end{aligned}
\end{equation}

%latex.default(resVar50, rowlabel = "", file = "resVar50.tex")%
\begin{table*}[!h]
	%\tiny	
	\begin{center}
		\begin{tabular}{llllllllll}
			\hline\hline
			\multicolumn{10}{c}{Number of variables $p=50$ } \tabularnewline
			\hline
			\multicolumn{1}{l}{}&\multicolumn{1}{c}{G1-S1}&\multicolumn{1}{c}{G1-S2}&\multicolumn{1}{c}{Gennet}&\multicolumn{1}{c}{nlasso}&\multicolumn{1}{c}{RBN}&\multicolumn{1}{c}{nSVM-L}&\multicolumn{1}{c}{nSVM-R}&\multicolumn{1}{c}{nSVM-S}&\multicolumn{1}{c}{nSVM-P}\tabularnewline
			\hline
			
			\rowcolor{LRed} TPR&$   0.51$&$   0.47$&$   0.11$&$   0.70$&$   0.65$&$   0.71$&$   0.69$&$   0.73$&$   \textbf{0.86}$\tabularnewline
			FPR&$   0.15$&$   0.11$&$   \textbf{0.01}$&$   0.17$&$   0.17$&$   0.22$&$   0.25$&$   0.33$&$   0.68$\tabularnewline
			\rowcolor{LRed} TNR&$   0.85$&$   0.89$&$   \textbf{0.99}$&$   0.83$&$   0.83$&$   0.78$&$  0.75$&$   0.67$&$   0.32$\tabularnewline
			FNR&$   0.49$&$   0.53$&$   0.89$&$   0.30$&$   0.35$&$   0.29$&$   0.31$&$   0.27$&$   \textbf{0.14}$\tabularnewline
			\rowcolor{LRed} MCE&$   0.17$&$   0.13$&$   \textbf{0.05}$&$ 0.18$&$  0.18$&$   0.23$&$ 0.25$&$   0.33$&$   0.66$\tabularnewline
			
			\hline
			
			\multicolumn{10}{c}{Number of variables $p=100$ }\tabularnewline
			\hline
			\rowcolor{LRed} TPR&$   0.26$&$   0.21$&$   0.12$&$   0.52$&$   0.39$&$  0.53$&$   0.53$&$   0.55$&$   \textbf{0.78}$\tabularnewline
			FPR&$   0.11$&$   0.07$&$   \textbf{0.01}$&$   0.12$&$   0.13$&$   0.18$&$   0.20$&$   0.22$&$   0.72$\tabularnewline
			\rowcolor{LRed} TNR&$   0.89$&$   0.93$&$  \textbf{0.99}$&$   0.88$&$   0.87$&$   0.82$&$ 0.80$&$   0.78$&$   0.28$\tabularnewline
			FNR&$   0.74$&$   0.79$&$   0.88$&$   0.48$&$   0.61$&$   0.47$&$   0.47$&$   0.45$&$   \textbf{0.22}$\tabularnewline
			\rowcolor{LRed} MCE&$   0.14$&$   0.11$&$   \textbf{0.06}$&$   0.14$&$   0.16$&$  0.19$&$  0.22$&$  0.23$&$   0.69$\tabularnewline
			\hline
		\end{tabular}
	\end{center}
		\caption{Linear simulated data, p=50, 100, n=20, pi=0.05, the last four columns correspond to the neighborhood SVM approach (nSVM)  using different kernels (L: Linear, R: Radial, S: Sigmod, P: Polynomial). }
	\label{tab:VAR}
\end{table*}

Table \ref{tab:VAR}, \ref{tab:nonVAR} and \ref{tab:VARnonVAR} present the results for the linear, nonlinear and mixture cases respectively for two values of p ($p=50$, $p=100$) and $n$ being fixed to 20.

As expected in all cases the performances decrease for high dimensions ($p=100$)
and quite good for all the methods in the linear case. The worst performance for all the methods is observed in the nonlinear case. Given the low rate (5\%) of edges present in the true network, the hardest task is to retrieve these edges thus to get high TPRs. High values of TNR are quite easy to achieve and correspond systematically to low values of the misclassification error  (MCE) rates.

The nSVM is the only approach where  TPR is above $50\%$.  As there is no global index to measure the fair performances of these approaches we try in general to have a good trade-off between TPR and TNR.

Figures \ref{fig:VAR}, \ref{fig:nonVAR}, and \ref{fig:VARnonVAR} show the position of the approaches we have compared in the space (TPR-TNR).

For the linear results, table \ref{tab:VAR} shows that the average number of edges which are correctly included into the estimate of the edges set is high in nSVM-S, nlasso, nSVM-L, nSVM-R and RBN respectively when $p=50$. These highly average values correspond also to high average values of edges which are correctly not included into the estimate of the edges set and to low average values of MCE, this is obvious from points in the upper right-hand side square in figure \ref{fig:VAR} with red color. 

When $p=100$, the average number of edges which are correctly included and not include into the estimate of the edges set breaks down in  RBN approach and be out of performance, but still significant in nSVM-S, nSVM-L, nSVM-R and nlasso respectively, See the blue points in the upper right hand side square in figure \ref{fig:VAR}.

In nonlinear simulation, nSVM still gives the highly significant results in both cases when  $p=50$ or $p=100$. Note that the average number of edges which are correctly included and not included into the estimate of the edge set when $p=50$ is high and closed its corresponds one when $p=100$ in nSVM with sigmod kernel which shows the stability of the results that are a function of the number of variables $p$. See the red and blue points in the upper right-hand side square in figure \ref{fig:nonVAR}.

In table \ref{tab:VARnonVAR} which is the simulated results of mixture linear and nonlinear time series data, nSVM  approach also gives highly significant results in both cases of different number of variables, especially in nSVM-S. On the other side, RBN and G1-S1 break down when the number of variables increases.  See the red and blue points in the upper right-hand side square in figure \ref{fig:VARnonVAR}.

In all the simulations the MCE decreases as the number of variables increase.
Moreover, nSVM approach especially SVM-S shows the best performance in finding the most correctly edges that include into the estimate of the edges set. RBN  sensitives to the linearity assumption and the number of variables especially when it is higher than the number of instances, this due to the likelihood function that used to estimate the network.

%latex.default(resNonVar50, rowlabel = "", file = "resNonVar50.tex")%
\begin{table*}[!h]
	%\tiny	
	\begin{center}
		\begin{tabular}{llllllllll}
			\hline\hline
			\multicolumn{10}{c}{Number of variables $p=50$ } \tabularnewline
			\hline
			\multicolumn{1}{l}{}&\multicolumn{1}{c}{G1-S1}&\multicolumn{1}{c}{G1-S2}&\multicolumn{1}{c}{Gennet}&\multicolumn{1}{c}{nlasso}&\multicolumn{1}{c}{RBN}&\multicolumn{1}{c}{nSVM-L}&\multicolumn{1}{c}{nSVM-R}&\multicolumn{1}{c}{nSVM-S}&\multicolumn{1}{c}{nSVM-P}\tabularnewline
			\hline
			
			\rowcolor{LRed} TPR&$   0.49$&$   0.39$&$   0.24$&$   0.31$&$   0.47$&$   0.56$&$   0.51$&$   0.55$&$   \textbf{0.66}$\tabularnewline
			FPR&$   0.22$&$   0.16$&$   \textbf{0.04}$&$   0.10$&$   0.31$&$   0.35$&$   0.28$&$   0.34$&$   0.52$\tabularnewline
			\rowcolor{LRed} TNR&$   0.78$&$   0.84$&$   \textbf{0.96}$&$   0.90$&$   0.69$&$  0.65$&$   0.72$&$  0.66$&$   0.48$\tabularnewline
			FNR&$   0.51$&$   0.61$&$   0.76$&$   0.69$&$   0.53$&$   0.44$&$   0.49$&$   0.45$&$   \textbf{0.34}$\tabularnewline
			\rowcolor{LRed} MCE&$   0.24$&$   0.18$&$   \textbf{0.07}$&$   0.13$&$   0.32$&$   0.35$&$ 0.29$&$   0.35$&$   0.51$\tabularnewline
			\hline
			
			\multicolumn{10}{c}{Number of variables $p=100$ } \tabularnewline
			\hline

			\rowcolor{LRed} TPR&$   0.43$&$   0.29$&$   0.17$&$   0.29$&$   0.32$&$   0.49$&$   0.48$&$  0.53$&$   \textbf{0.67}$\tabularnewline
			FPR&$   0.17$&$   0.07$&$   \textbf{0.03}$&$   0.07$&$   0.16$&$   0.23$&$   0.22$&$   0.30$&$   0.52$\tabularnewline
			\rowcolor{LRed} TNR&$   0.83$&$   0.93$&$   \textbf{0.97}$&$   0.93$&$   0.84$&$   0.77$&$   0.78$&$  0.70$&$   0.48$\tabularnewline
			FNR&$   0.57$&$   0.71$&$   0.83$&$   0.71$&$   0.68$&$   0.51$&$   0.52$&$   0.47$&$   \textbf{0.33}$\tabularnewline
			\rowcolor{LRed} MCE&$   0.19$&$   0.10$&$   \textbf{0.07}$&$   0.10$&$   0.19$&$   0.24$&$   0.24$&$   0.31$&$   0.52$\tabularnewline
			
			\hline
		\end{tabular}
	\end{center}
		\caption{Noninear simulated data, p=50, 100, n=20, pi=0.05, the last four columns correspond to the neighborhood SVM approach (nSVM)  using different kernels (L: Linear, R: Radial, S: Sigmod, P: Polynomial).}
	\label{tab:nonVAR}
\end{table*}

%latex.default(resVarNonVar50, rowlabel = "", file = "resVarNonVar50.tex")%
\begin{table*}[!h]
	%\tiny	
	\begin{center}
		\begin{tabular}{llllllllll}
			\hline\hline
			\multicolumn{10}{c}{Number of variables $p=50$ } \tabularnewline
			\hline
			\multicolumn{1}{l}{}&\multicolumn{1}{c}{G1-S1}&\multicolumn{1}{c}{G1-S2}&\multicolumn{1}{c}{Gennet}&\multicolumn{1}{c}{nlasso}&\multicolumn{1}{c}{RBN}&\multicolumn{1}{c}{nSVM-L}&\multicolumn{1}{c}{nSVM-R}&\multicolumn{1}{c}{nSVM-S}&\multicolumn{1}{c}{nSVM-P}\tabularnewline
			\hline

			\rowcolor{LRed} TPR&$  0.61$&$   0.49$&$   0.33$&$   0.42$&$   0.54$&$   0.66$&$  0.60$&$  0.64$&$   \textbf{0.76}$\tabularnewline
			FPR&$   0.24$&$   0.16$&$   \textbf{0.03}$&$   0.11$&$   0.32$&$   0.34$&$   0.27$&$   0.33$&$   0.54$\tabularnewline
			\rowcolor{LRed} TNR&$  0.76$&$   0.84$&$   \textbf{0.97}$&$   0.89$&$   0.68$&$   0.66$&$ 0.73$&$ 0.67$&$   0.46$\tabularnewline
			FNR&$   0.39$&$   0.51$&$   0.67$&$   0.58$&$   0.46$&$   0.34$&$   0.40$&$   0.36$&$   \textbf{0.24}$\tabularnewline
			\rowcolor{LRed} MCE&$   0.25$&$   0.18$&$   \textbf{0.07}$&$   0.13$&$   0.33$&$   0.34$&$  0.28$&$   0.34$&$   0.53$\tabularnewline
			\hline
			
			\multicolumn{10}{c}{Number of variables $p=100$ } \tabularnewline
			\hline
			
			\rowcolor{LRed} TPR&$   0.46$&$   0.31$&$   0.14$&$   0.34$&$   0.34$&$   0.53$&$  0.51$&$ 0.56$&$   \textbf{0.66}$\tabularnewline
			FPR&$   0.17$&$   0.07$&$   \textbf{0.02}$&$   0.07$&$   0.16$&$   0.22$&$   0.22$&$   0.29$&$   0.49$\tabularnewline
			\rowcolor{LRed} TNR&$   0.83$&$   0.93$&$   \textbf{0.98}$&$   0.93$&$   0.84$&$  0.78$&$ 0.78$&$ 0.71$&$   0.51$\tabularnewline
			FNR&$   0.54$&$   0.69$&$   0.86$&$   0.66$&$   0.66$&$   0.47$&$   0.49$&$   0.44$&$   \textbf{0.34}$\tabularnewline
			\rowcolor{LRed} MCE&$   0.19$&$   0.10$&$   \textbf{0.06}$&$   0.10$&$   0.19$&$  0.24$&$  0.23$&$ 0.30$&$   0.48$\tabularnewline
			
			\hline
		\end{tabular}
	\end{center}
		\caption{Linear and nonlinear simulated data, p=50, 100, n=20, pi=0.05, the last four columns correspond to the neighborhood SVM approach (nSVM)  using different kernels (L: Linear, R: Radial, S: Sigmod, P: Polynomial).}
	\label{tab:VARnonVAR}
\end{table*}

\section{Conclusion}
SVM criterion is an efficient approach to find the interaction points to approximate the structure of the data sets. The ability to use different types of kernels allow SVMs to find the best active sets that construct the model according to the types of the input data. Also, SVM is not sensitive too much to the number of variables inserted as we shown in nSVM-S. Further work is to apply these results to gene expression data and compare it with approximate true structures.
\section*{Acknowledgment}
The authors are grateful to "HERMES" project,  Erasmus Mundus European programme, action 2,  Marseilles, France, for the financial support during his study.

\bibliographystyle{splncs04}
\bibliography{mainTex}

\end{document}